\documentclass[letterpaper]{article} % DO NOT CHANGE THIS
\usepackage{aaai25}  % DO NOT CHANGE THIS
\usepackage{times}  % DO NOT CHANGE THIS
\usepackage{helvet}  % DO NOT CHANGE THIS
\usepackage{courier}  % DO NOT CHANGE THIS
\usepackage[hyphens]{url}  % DO NOT CHANGE THIS
\usepackage{graphicx} % DO NOT CHANGE THIS
\usepackage{natbib}  % DO NOT CHANGE THIS AND DO NOT ADD ANY OPTIONS TO IT
\usepackage{caption} % DO NOT CHANGE THIS AND DO NOT ADD ANY OPTIONS TO IT
\usepackage{algorithm}
\usepackage{algorithmic}
\usepackage{amsmath}
\usepackage{multirow}

\usepackage{newfloat}
\usepackage{listings}
\DeclareCaptionStyle{ruled}{labelfont=normalfont,labelsep=colon,strut=off} % DO NOT CHANGE THIS
\floatstyle{ruled}
\newfloat{listing}{tb}{lst}{}
\floatname{listing}{Listing}
\title{A Sample-Level Evaluation and Generative Framework for Model Inversion Attacks}
\author{
    Haoyang Li\textsuperscript{\rm 1},
    Li Bai\textsuperscript{\rm 1},
    Qingqing Ye\textsuperscript{\rm 1},
    Haibo Hu\textsuperscript{\rm 1}\thanks{Corresponding author},
    Yaxin Xiao\textsuperscript{\rm 1},
    Huadi Zheng\textsuperscript{\rm 2},
    Jianliang Xu\textsuperscript{\rm 3}
}
\affiliations{
    \textsuperscript{\rm 1}Department of Electrical and Electronic Engineering, The Hong Kong Polytechnic University \\
    \textsuperscript{\rm 2}Huawei Technology \\
    \textsuperscript{\rm 3} Department of Computer Science, Hong Kong Baptist University \\
    \{hao-yang9905.li, baili.bai\}@connect.polyu.hk, \{qqing.ye, haibo.hu\}@polyu.edu.hk, 20034165r@connect.polyu.hk, zhenghuadi@huawei.com, xujl@comp.hkbu.edu.hk
}

\usepackage{bibentry}
\begin{document}

\maketitle
\begin{abstract}
Model Inversion (MI) attacks, which reconstruct the training dataset of neural networks, pose significant privacy concerns in machine learning. Recent MI attacks have managed to reconstruct realistic label-level private data, such as the general appearance of a target person from all training images labeled on him. Beyond label-level privacy, in this paper we show sample-level privacy, the private information of a single target sample, is also important but under-explored in the MI literature due to the limitations of existing evaluation metrics. To address this gap, this study introduces a novel metric tailored for training-sample analysis, namely, the Diversity and Distance Composite Score (DDCS), which evaluates the reconstruction fidelity of each training sample by encompassing various MI attack attributes. This, in turn, enhances the precision of sample-level privacy assessments.

Leveraging DDCS as a new evaluative lens, we observe that many training samples remain resilient against even the most advanced MI attack. As such, we further propose a transfer learning framework that augments the generative capabilities of MI attackers through the integration of entropy loss and natural gradient descent. Extensive experiments verify the effectiveness of our framework on improving state-of-the-art MI attacks over various metrics including DDCS, coverage and FID. Finally, we demonstrate that DDCS can also be useful for MI defense, by identifying samples susceptible to MI attacks in an unsupervised manner.
\end{abstract} 

\section{Introduction}
With today's explosive growth in Machine Learning (ML), all businesses are collecting personal data more frequently than ever~\cite{astaple3}.
As such, there is an increasing demand of privacy preservation during model training and deployment~\cite{astaple1, astaple2, astaple4, astaple5}.
However, it is well known that by exploiting the confidence score of neural networks, Model Inversion (MI) attacks can partially reconstruct the training dataset from an ML model~\cite{mi15}.
Compared with membership or property inference attacks that only infer certain membership or property information as a binary classification task~\cite{mif22sp}, an MI attack targets at the whole private information in the training dataset and reconstructs multiple private features simultaneously~\cite{misurvey22}, which poses even more severe threats than its inference counterpart.
In line with previous studies~\cite{gmi20cvpr, kedmi21iccv, ppa22icml}, we consider white-box MI attacks, where the attacker has access to the specific parameters and architecture of the victim model.

Given the inherent difficulty for MI attacks to recover every sample within a large-scale dataset, recent works~\cite{ppa22icml,mirror22ndss} have lowered down their goal to recover one or few representatives in each class label. This trend was first motivated in the field of face recognition~\cite{faceevolve21}, where each label corresponds to the identity of a person. Since these label representatives can reveal the general appearance of a person, they arise as a significant concern in recent studies~\cite{vmi21nips,rethink23cvpr}. However, the potential for privacy compromise varies across applications. In some domains beyond facial recognition, label representatives may convey little or even no private information, which in turn weakens such threats. For instance, in disease diagnosis, each label corresponds to a well-known disease, so label representatives merely encapsulate public knowledge about the conditions of this disease. However, MI attacks may still threaten {\bf sample-level} privacy by reconstructing individual samples.

To address the imperative to unearth and safeguard sample-level privacy, this study critiques existing evaluation metrics of MI by first demonstrating their inadequacy in fully capturing the multifaceted characteristics of MI attacks.
Furthermore, metrics based on distributional similarity are sensitive to the distribution of reconstructed samples, so that their evaluation results can be easily fooled by controlling the sample distribution (\emph{e.g.}, producing redundant samples or approximating the distribution with fewer samples).
To overcome these limitations, we introduce a novel metric, namely \textit{Diversity and Distance Compostie Score} (DDCS), that assesses the integrity of sample-level privacy in the context of MI attacks.
In essence, DDCS uniquely evaluates the extent of reconstruction by the attack for each sample in the training dataset, so that it can comprehensively incorporate diverse attributes (\emph{e.g.}, distance, coverage and distributional similarity) to indicate a good MI attack and is robust against sample distribution manipulations.
From the defense's perspective, DDCS quantifies the susceptibility of individual samples to MI attacks by assigning each with a reconstruction distance, enabling the identification and protection of vulnerable samples in an unsupervised manner.

Leveraging Diversity and Distance Composite Score (DDCS) to evaluate the success of reconstruction, we observe that many samples in the training dataset remain un-reconstructable by even the most advanced MI attacks~\cite{ppa22icml}.
This observation coincides with the limited diversity and coverage of current MI methodologies, mainly because prior metrics focus primarily on label-level privacy.
In response, we introduce a Generative Adversarial Network (GAN) augmentation framework designed to enhance both coverage and diversity of MI attacks, through mitigating the produced artifacts while reserving the advantages from the inversion-specific GAN~\cite{stylegan2_20cvpr, kedmi21iccv}.
To achieve this, our framework transfers a pre-trained GAN with entropy loss regularized with natural gradient descent~\cite{ngd}, so as to expand the attacker's generative ability and mitigate artifacts typically associated with deep GAN structures, such as those found in StyleGAN~\cite{stylegan2_20cvpr}.

To validate our methodologies, we conduct comprehensive experiments on face recognition and dog breed classification datasets under various model architectures and MI algorithms. The results affirm the comprehensiveness of DDCS, underscoring its utility in evaluating sample-level privacy and robustness against manipulations of sample distributions. Further, the experiments substantiate the efficacy of our GAN augmentation framework in enhancing the performance of leading MI attacks, as evidenced by improvements in Fr\'echet Inception Distance, coverage, and DDCS metrics. Additionally, we also conduct a series of ablation studies to scrutinize the influence of different experimental parameters, ensuring a thorough understanding of our framework's dynamics and its implications for privacy-preserving ML.
To summarize, our contributions in this paper are as follows:
\begin{itemize}
	\item
	We introduce DDCS, a comprehensive evaluation metric for MI attacks with great potential for sample-level privacy analysis.
	\item
	We propose a GAN augmentation framework to improve MI attacks on DDCS while retaining the image quality.
	\item
	We conduct extensive experiments on computer vision tasks to demonstrate DDCS and verify the effectiveness our framework.
\end{itemize}

The rest of the paper is organized as follows.
We first review recent work on MI attacks.
Later, we point out the limitations of evaluation metrics and introduce DDCS.
We then propose our GAN augmentation framework and conduct extensive experiments before the conclusion. 

%The rest of the paper is organized as follows.
%In Section~\ref{related} we review recent work on MI attacks.
%We point out the limitations of evaluation metrics in Section~\ref{sec:ddcs:limitation} and introduce DDCS in Section~\ref{sec:ddcs:intro}.
%We propose our GAN augmentation framework in Section~\ref{sec:gan_aug} and conduct extensive experiments in Section~\ref{sec:exp} with the conclusion in Section~\ref{sec:conclusion}. 
%\input{sec/2_related}
\section{DDCS: A New Evaluation Metric for MI}
Prior studies on MI attacks have employed a range of evaluation metrics, each with its own limitations.
Notably, many of these metrics overlook critical attributes of the reconstructed data, such as diversity, and are susceptible to alteration in data distribution caused by non-essential factors, such as the presence of redundant samples.
This issue is prevalent across MI attack evaluations.
In the rest of this section, we delve into the impact of these limitations on the thorough assessment of MI attacks and introduce a novel metric designed to overcome these challenges.

\subsection{Limitations of Existing Metrics}
\label{sec:ddcs:limitation}
An ideal MI attack~\cite{vmi21nips}, as the name suggests, aims to accurately restore the target training dataset $\mathcal{D}_{tar}$. As such, attacks which can successfully reconstruct every sample in $\mathcal{D}_{tar}$ are more severe than those only recover sensitive labels.
While recent studies have highlighted the risks of label-level privacy breaches in tasks like face recognition, where label representatives can disclose an individual's identity and appearance~\cite{mirror22ndss}, we argue that the reconstruction of sample-level data poses a far greater threat to privacy.
For example, in face recognition, the variety in reconstructed samples can expose more nuanced personal details, such as an individual's preferred hairstyles, accessories, and expressions~\cite{vmi21nips}.
This is particularly true in fields beyond face recognition, such as disease diagnosis and dog breed classification, where labels convey commonly known information.

The oversight of sample-level privacy may stem from the inadequacy of existing evaluation metrics to fully and precisely evaluate crucial characteristics indicative of successful MI attacks. To address this gap, we draw three fundamental traits that are essential for an ideal MI attack on sample-level privacy --- accuracy/distance, coverage, and distributional similarity. This discussion also explores the shortcomings of current evaluation methodologies in accurately assessing these critical attributes.

\noindent
\textbf{Accuracy/Distance.}\quad
Accuracy and distance both encourage MI attacks focusing on label-level privacy to generates label representatives.
Recall that the objective function of MI is to maximize the sample's output confidence given the victim model~\cite{mi15}.
Thus, the optimization of MI attacks can easily overfit to the specific structure of the victim neural network, and a successful MI attack should then be independent of the classifier architecture.
Therefore, previous works train a model with a different structure from the victim model and measure the accuracy of reconstructed dataset $\mathcal{D}_{rec}$.
In this context, high classification accuracy on the reconstructed dataset $\mathcal{D}_{rec}$ signifies a reduced distance between $\mathcal{D}_{rec}$ and the target dataset $\mathcal{D}_{tar}$.
Comparable to accuracy, feature distance is another metric frequently employed in MI assessments, quantifying the $l_2$ distance between reconstructed and target samples within the evaluator's feature space.

\begin{figure}[htb]%[htbp]
	\centering
	\includegraphics[width=0.9\linewidth]{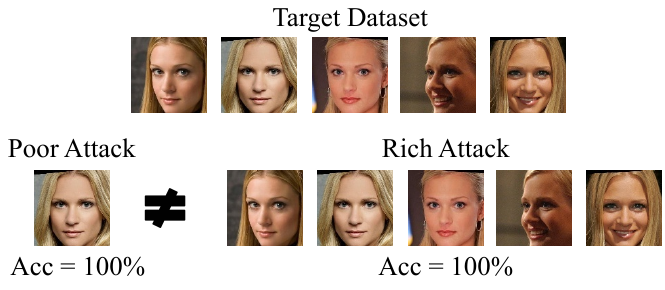}
	\caption{When each reconstructed sample is close enough to the target dataset, average accuracy or distance fail to differentiate between the rich and poor attack.}
	\label{fig:toy_accuracy}
\end{figure}

\noindent
\textbf{Coverage.}\quad Coverage refers to the proportion of $\mathcal{D}_{tar}$ that are reconstructed by $\mathcal{D}_{rec}$ and can be increased by improving the diversity of $\mathcal{D}_{rec}$. While accuracy and feature distance are the primary metrics for evaluating MI attacks, they overlook the diversity of samples and fail to comprehensively assess the coverage of MI attacks.
In the toy example of Figure~\ref{fig:toy_accuracy} where $\mathcal{D}_{tar}$ only contains five images, there is a rich attack that can successfully recover all images in $\mathcal{D}_{tar}$ and a poor attack that recovers only one of them.
If all samples from the rich and poor attack are identical to some images in $\mathcal{D}_{rec}$, both attacks achieve the same average accuracy, albeit the rich attack is obviously more threatful than the poor attack.

\begin{figure}[htb]%[htbp]
	\centering
	\includegraphics[width=0.9\linewidth]{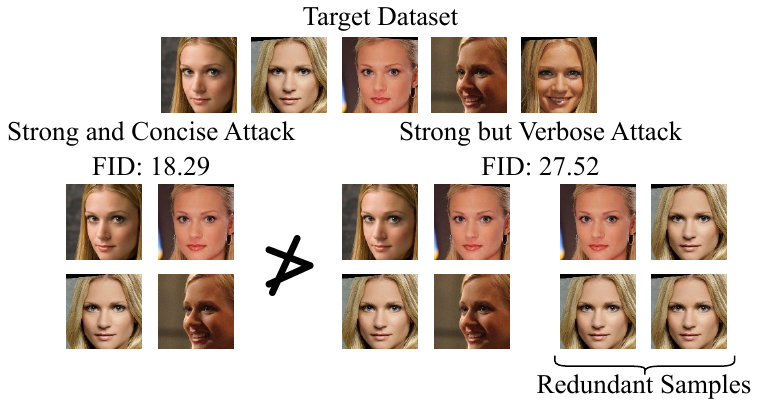}
  \vspace{-0.1in}
	\caption{Redundant samples change the data distribution, thus affecting the FID calculation, although both attacks successfully recover four images from the target dataset.}
	\label{fig:toy_fid}
 \vspace{-0.2in}
\end{figure}

\noindent
\textbf{Distributional Similarity.}\quad
Being aware of the limitation in accuracy and distance, recent works~\cite{kedmi21iccv,vmi21nips,ppa22icml} have incorporated distributional similarity, \emph{e.g.}, Fr\'echet Inception Distance (FID)~\cite{fid}, as an additional metric by treating $\mathcal{D}_{tar}$ and $\mathcal{D}_{rec}$ as two distinct image distributions.
Since the diversity of $\mathcal{D}_{rec}$ will affect its distribution, the problem of Figure~\ref{fig:toy_accuracy} can be alleviated with distributional similarity.
However, the distribution of images can be easily affected by factors that are trivial in privacy leakage, leading to misjudgment of MI attacks.
One example is the bias caused by redundant samples that MI attacks can easily generate, which leads to a distribution shift between $\mathcal{D}_{tar}$ and $\mathcal{D}_{rec}$, thereby causing a worse FID that is not true.
As in Figure~\ref{fig:toy_fid}, there is an attack that can restore most of the samples without any duplication (\emph{i.e.}, strong and concise attack), and an attack that can restore the same diverse samples with some duplications (\emph{i.e.}, strong but verbose attack).
The former attack can achieve a better FID, but it is premature to conclude that the former attack is more powerful than the latter, because they both successfully restore four images in $\mathcal{D}_{tar}$.
This phenomenon arises because distributions can be approximated using varying sample batches, allowing for the manipulation of metrics by altering the sample distribution. We will empirically show how FID is susceptible to such manipulations in the experiment, and therefore these factors do not fundamentally characterize a good MI attack.

\subsection{Diversity and Distance Composite Score}
\label{sec:ddcs:intro}
We believe the core reason for the above limitations is that these metrics overly concentrate on the distribution of reconstructed samples ($\mathcal{D}_{rec}$-oriented), which leads to an incomplete assessment of privacy leakage in the target samples $\mathcal{D}_{tar}$.
This approach diverges from the objectives of ideal MI attacks, as outlined previously.
Consequently, an MI algorithm could generate high-fidelity label representatives and manipulate the distribution of $\mathcal{D}_{rec}$ to perform favorably according to these metrics.

Different from $\mathcal{D}_{rec}$-oriented metrics introduced in previous works, we truly measure the degree to which $\mathcal{D}_{tar}$ is recovered by $\mathcal{D}_{rec}$, and propose a more robust and comprehensive evaluation metric, namely \textit{Diversity and Distance Composite Score} (DDCS).
We start from the ideal MI attack, where each sample from $\mathcal{D}_{tar}$ has its one-to-one matching in $\mathcal{D}_{rec}$ (\emph{i.e.}, this attack achieves perfect diversity and distance regarding $\mathcal{D}_{tar}$), and any other attack that reduces $\mathcal{D}_{rec}$'s diversity and distance to $\mathcal{D}_{tar}$ is a suboptimal attack, which inspires us how to calculate DDCS, as shown in Algorithm~\ref{alg:ddcs}.
One worth-noting term used in DDCS is the \textit{Reconstruction Distance} $d^i_{tar}$ for every target sample $x^i_{tar} \in \mathcal{D}_{tar}$.
Specifically, given a distance metric $d(\cdot,\cdot)$, if $x^j_{rec}$'s closest target sample is $x^i_{tar}$ with the distance as $d(x^j_{rec}, x^i_{tar})$, $x^i_{tar}$ is said to be reconstructed by $x^j_{rec}$ with a reconstruction distance of $d(x^j_{rec}, x^i_{tar})$ (Line~\ref{alg:rec_dist}).
$d(x^j_{rec}, x^i_{tar})$ is then recorded into a set, denoted by $S_{tar}^{i}$ of $x^i_{tar}$ for further handling.

\begin{algorithm}[htb]
	\caption{DDCS Computation}
	\label{alg:ddcs}
    \textbf{Input}: Target dataset $\mathcal{D}_{tar}$, reconstructed dataset    $\mathcal{D}_{rec}$, constant $c$ \\
    \textbf{Output}: DDCS$_{\text{avg}}$, DDCS$_{\text{best}}$ %for $\mathcal{D}_{rec}$
	\begin{algorithmic}[1]
	\STATE Initialize $S_{tar}^i=\emptyset$ for $x^i_{tar} \in \mathcal{D}_{tar}$
	
	\FOR {$x^j_{rec} \in \mathcal{D}_{rec}$}
    \STATE Record $d_{tar}^i=d(x^i_{tar},x^j_{rec})$ into $S_{tar}^i$, where $x^{j}_{rec}$ and $x^{i}_{tar}$ achieve the smallest $d_{tar}^i$ for $x^{i}_{tar} \in \mathcal{D}_{tar}$ \label{alg:rec_dist}
	\ENDFOR
	
	\STATE Initialize DDCS$_{\text{avg}}=0$, DDCS$_{\text{best}}=0$
	
	\FOR {$x^i_{tar} \in \mathcal{D}_{tar}$}
    \STATE DDCS$_{\text{avg}} += mdf[\text{avg}(S_{tar}^i)+c]$ \label{alg:ddcs:avg}
	\STATE DDCS$_{\text{best}} += mdf[\text{min}(S_{tar}^i)+c]$ \label{alg:ddcs:best}
    \ENDFOR
	
	\STATE Normalize DDCS$_{\text{avg}}$ and DDCS$_{\text{best}}$ with $|\mathcal{D}_{tar}|$ 
 \end{algorithmic}
 \label{alg:ddcs:normalize}
\end{algorithm}

In Lines~\ref{alg:ddcs:avg} and~\ref{alg:ddcs:best}, we propose two variants of DDCS tailored to distinct scenarios.
To ensure that higher DDCS values indicate better outcomes, we employ monotonically decreasing functions, $mdf[\cdot]$, in its calculation, specifically utilizing the simple reciprocal function on $\text{avg}(S_{tar}^i)+c$ and $\text{min}(S_{tar}^i)+c$ in this paper.
DDCS$_{\text{avg}}$ quantifies the overall privacy risk of $\mathcal{D}_{tar}$ against $\mathcal{D}_{rec}$, calculated as the sum of the reciprocals of the averages for each $S^i_{tar}$, \emph{i.e.}, $\text{avg}(S_{tar}^i)$.
DDCS$_{\text{best}}$ assesses the minimum reconstruction distance across all $S^i_{tar}$, \emph{i.e.}, $\text{min}(S_{tar}^i)$, offering insight into the potential maximum effectiveness of an attack on $\mathcal{D}_{tar}$.
If $x^i_{tar}$ is not recovered by any reconstructed sample, we set their $mdf[\text{avg}(S^i_{tar}) + c]$ and $mdf[\text{min}(S^i_{tar}) + c]$ to 0.
Besides, we reserve a constance $c$ to regularize the range of DDCS when $d^i_{tar}$ is so small.
As a final note, both DDCS$_{\text{avg}}$ and DDCS$_{\text{best}}$ are normalized by the total number of samples $|\mathcal{D}_{tar}|$ in the target dataset (Line~\ref{alg:ddcs:normalize}).

DDCS mitigates those limitations discussed before from two aspects.
Firstly, this metric addresses the privacy for {\bf every} sample in the target dataset ($\mathcal{D}_{tar}$-oriented).
Thus, DDCS encourages MI attacks to focus on sample-level privacy instead of label representatives, with high robustness against distribution change of $\mathcal{D}_{rec}$ as in Figure~\ref{fig:toy_fid}.
Secondly, DDCS takes multiple important characteristics into account, so that an MI attack can win this metric by truly attacking the privacy of target samples.
For example, if an attack reaches higher coverage, more samples from $\mathcal{D}_{tar}$ will be reconstructed, increasing DDCS value and mitigating the problem in Figure~\ref{fig:toy_accuracy}.
On the other hand, when diversity remains unchanged, producing better reconstructed sample to reduce the reconstruction distance can also improve DDCS. Besides, reconstruction distance of DDCS can also be harnessed beyond attacks, such as measuring per-example vulnerability for $\mathcal{D}_{tar}$ to selectively protect those vulnerable samples, which will be discussed in detail in the experiment.
This application scenario  has been recently studied in the context of membership inference attacks~\cite{mif22sp}, but we are the first to study it in MI scenarios.

\section{Enhancing MI Attacks over DDCS}
\label{sec:gan_aug}
\begin{figure}[htb]%[htbp]
	\centering
 %\vspace{-0.1in}
	\includegraphics[width=0.9\linewidth]{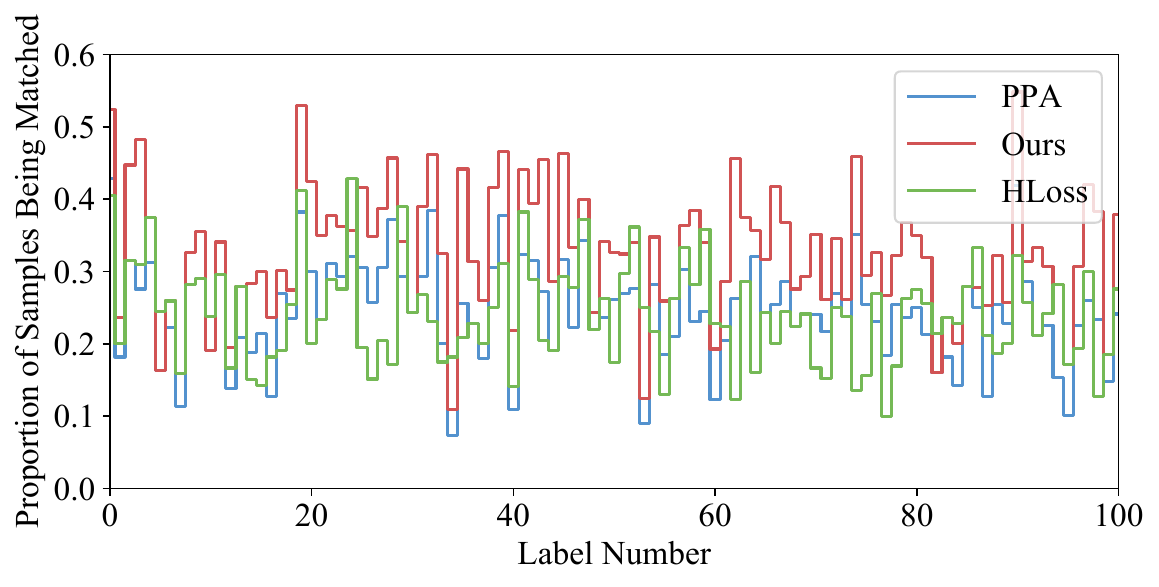}
 \vspace{-0.1in}
	\caption{Visualization of DDCS indicating the proportion of samples for the first 100 labels of VGG16BN-UMDFaces that are matched to $\mathcal{D}_{rec}$ with three attacks (PPA, HLoss and Ours). As explained in the section of DDCS, a target sample $x^i_{tar}$ will be matched if its set $S^i_{tar}$ is non-empty after the execution of Algorithm~\ref{alg:ddcs}.}
	\label{fig:ddcs_coverage}
 %\vspace{-0.1in}
\end{figure}

Based on the $\mathcal{D}_{tar}$-oriented evaluation by DDCS, we evaluate the actual extent of privacy intrusion by attackers into $\mathcal{D}_{tar}$, as depicted in Figure~\ref{fig:ddcs_coverage}.
This figure illustrates the percentage of per-label samples in $\mathcal{D}_{tar}$ that are reconstructable by at least one counterpart in $\mathcal{D}_{rec}$, reflecting the attack's coverage.
Despite the focus of prior research on producing high-quality label representatives, we note that even the most advanced MI attacks~\cite{ppa22icml} fall short in achieving comprehensive coverage.
This revelation significantly undermines the perceived threat level of MI attacks, as the ML training party can easily identify these samples using DDCS and protect them.
This insight underscores the need for us to enhance the generative capabilities of attackers in existing MI attack methodologies.

\subsection{Difficulties on Training with Entropy Loss}
Enhancing the generative capacity of a GAN for MI attacks involves refining the training methodology of the generative network, allowing the generator $G$ to leverage knowledge from the victim model $V$. To facilitate more effective utilization of $V$'s knowledge by the attacker, KEDMI~\cite{kedmi21iccv} addresses this by incorporating an additional entropy loss $\mathcal{L}_{\text{adv}}$, into the standard DCGAN training regimen of $G$.
$\mathcal{L}_{\text{adv}}$ is calculated based on the information entropy of $V$'s confidence score for a set of generated images $G(z)$.
Minimizing $\mathcal{L}_{\text{adv}}$ aims to enhance the similarity between $G(z)$ and the private training datasets of $V$.
However, this approach introduces two notable challenges:

\noindent
\textbf{P1.}\quad
The addition of $\mathcal{L}_{\text{adv}}$ to the training deviates from the GAN's original objective of maximizing image quality, consequently compromising it.

\noindent
\textbf{P2.}\quad Training a GAN from the scratch with $\mathcal{L}_{\text{adv}}$ is redundant under the assumption of publicly available pre-trained GANs.

\begin{figure}[htb]%[htbp]
	\centering
	\includegraphics[width=0.95\linewidth]{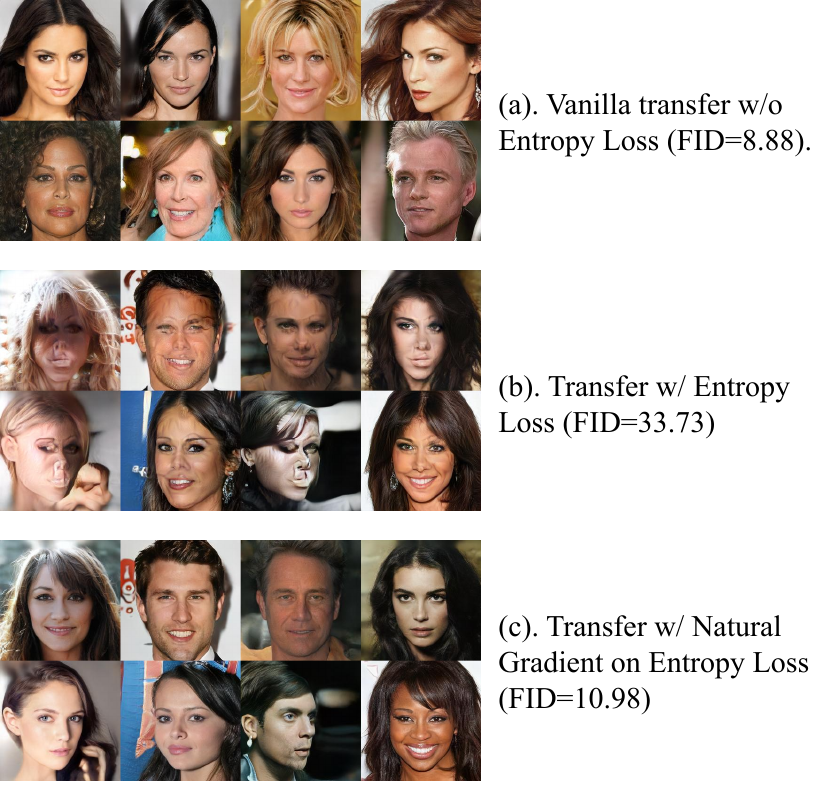}
	\caption{Snapshots and image quality of generated images for three different approaches. Image quality, evaluated by FIDs, are calculated with the same random seed and training configurations. Snapshots are generated using the same and fixed latent codes.}
	\label{fig:transfer_quality}
 \vspace{-0.1in}
\end{figure}

Compared to P2, P1 poses a more significant challenge, as evidenced by recent findings~\cite{ppa22icml} where the KEDMI approach underperforms traditional methods when integrated with StyleGAN.
We attribute P1 to the optimization of entropy loss under a loosely enforced image quality constraint, which inadvertently leads to the generation of artifacts rather than the intended private information of $V$.
This issue is exemplified in Figures~\ref{fig:transfer_quality} (a) and (b), where, despite the intention for entropy loss to enhance $V$'s confidence in the generated images, it instead prompts $G$ to produce artifacts rather than accurate representations of target dataset $\mathcal{D}_{tar}$.
In this figure, FID assesses image quality, different from its use in MI attacks where it evaluates attack performance.

\subsection{Natural Gradient for Entropy Loss}
\label{sec:gan_aug:ngd}
Our methodology for refining entropy loss draws upon the manifold hypothesis, which states that the distribution of natural images, considered as a high-dimensional manifold, is representable through latent codes on a lower-dimensional manifold~\cite{ml16goodfellow,rogan18cvpr}.
In other words, artifacts caused by entropy loss are interpreted as the result of generated images deviating from their native manifold.
Therefore, we propose to mitigate the emergence of artifacts by maintaining the generated images within their original manifold throughout the training of $\mathcal{L}_{\text{adv}}$-assisted GAN.

To tackle the above P1, we incorporate Natural Gradient Descent (NGD)~\cite{ngd} into our methodology, aiming to preserve the manifold structure of generated images while training with entropy loss.
NGD, an approximate second-order optimization technique, is conceptualized as the optimization on the Riemannian manifold~\cite{ro}.
Its implementation relies on the inverse of Fisher Information Matrix (FIM), denoted as $\mathcal{F}^{-1}$, an intrinsic distance metric that facilitates the preservation of variable on the manifold through successive optimization steps.
To approximate $\mathcal{F}$, \cite{rogan21iclr} suggests utilizing the Hessian of a squared LPIPS distance function $H_{G(z)}(d^2(G(z),G(z_0)))$ as the metric tensor for the manifold of deep generative image models.
This insight underpins our proposed NGD-based strategy for optimizing entropy loss:
\begin{equation}
	P(\frac{\partial \mathcal{L}_{\text{adv}}}{\partial G}) = H^{-1}_{G(z)}(d^2(G(z),G(z_0))) \frac{\partial \mathcal{L}_{\text{adv}}}{\partial G(z)} \cdot \frac{\partial G(z)}{\partial G}
	\label{eq:ngd}
\end{equation}
In each optimization step, we project the original gradient $\frac{\partial \mathcal{L}_{\text{adv}}}{\partial G}$ onto its natural gradient $P(\frac{\partial \mathcal{L}_{\text{adv}}}{\partial G})$ using the inverse Hessian matrix of the squared distance function $H^{-1}_{G(z)}(d^2(G(z),G(z_0)))$ for a given batch of generated images $G(z)$ and their entropy loss $\mathcal{L}_{\text{adv}}$.
%The detailed derivation of equation~\ref{eq:ngd} is in Appendix A.
It is worth noting that alternative methodologies for specifying the metric tensor of the image manifold, such as employing the Jacobian of the generator $G$~\cite{rogan18cvpr}, are also viable within this framework.

Figure~\ref{fig:transfer_quality} (c) illustrates the efficacy of Equation~\ref{eq:ngd} in artifact mitigation and improvement of image quality, where FID is reduced from 33.73 to 10.98 and the artifacts are enhanced with realistic features.

To reduce the computational cost in P2, we adopt a transfer learning approach for our training regimen.
Utilizing a pre-trained GAN, we apply our training framework on an auxiliary dataset for a limited number of epochs.
In line with~\cite{rogan21iclr}, we employ the Hessian Vector Product (HVP) to efficiently approximate the Hessian, pre-compute a batch of HVPs, and then use them throughout the training process.
Notably, the image domains produced by a GAN trained with entropy loss diverge from those generated by a conventional (vanilla) GAN, leading to a restricted and less diverse $\mathcal{D}_{rec}$ when using a single GAN model~\cite{ppa22icml,vmi21nips}.
To counteract this limitation, we propose to augment a vanilla GAN with an $\mathcal{L}_{\text{adv}}$-enhanced GAN for a more diverse $\mathcal{D}_{rec}$.

%In the Appendix, we provide a comprehensive overview of the proposed algorithm.
To summarize, given a pre-trained GAN and victim model, we calculate the standard StyleGAN training loss as well as the entropy loss at each training step.
The gradients from the loss of standard training and entropy loss will be combined to update the generator $G$, while the gradient from the entropy loss will first be projected using Equation~\ref{eq:ngd} before applying gradient descent.
Finally, the optimized GAN, trained with this methodology, will produce the reconstructed samples following the standard MI recovery strategy.
\section{Experiments}
\label{sec:exp}
\subsection{Experimental Setup}
\noindent
\textbf{Datasets.}\quad
We conduct experiments\footnote{https://github.com/haoyangliASTAPLE/DDCS\_MI.git} on four high-resolution image datasets for two image classification tasks, namely, face recognition and dog breed classification.
For face recognition, we use UMDFaces~\cite{umdfaces} face dataset for training victim models and CelebA-HQ~\cite{pgan18iclr,celeba} dataset as the attacker's auxiliary dataset.
For dog breed classification, Stanford Dogs~\cite{stanforddogs} dataset is selected as the training dataset of victim models and Tsinghua Dogs~\cite{tsinghuadogs20} for the auxiliary datasets.
Similar to previous works~\cite{kedmi21iccv, gmi20cvpr}, we take a subset of 1000 identities from UMDFaces that contains 50981 training samples and 3000 test samples as $\mathcal{D}_{tar}$ to train the victim models.
Following~\cite{ppa22icml}, we use all labels for Stanford Dogs and split them into 18780 training samples and 1800 test samples.
To prevent strong assumptions about attacks, we limit the attacker's sample size in the auxiliary dataset to 30000 for both tasks.

\noindent
\textbf{Models.}\quad
We train 4 types of victim models: (1) VGG16~\cite{vgg} with Batch Normalization~\cite{bn} (VGG16BN), (2) ResNet50~\cite{resnet}, (3) Improved ResNet50 with Squeeze-and-Excitation~\cite{se18cvpr} blocks (IR50-SE), and (4) AlexNet~\cite{alexnet}.
We resume pre-trained StyleGAN2~\cite{stylegan2_20cvpr} of resolution 256 $\times$ 256 trained on two datasets, CelebA-HQ for face identification task and LSUN-Dogs~\cite{lsun} for dog breed classification task.

\noindent
\textbf{Competing MI Attacks.}\quad
Our GAN augmentation framework (Ours) is compared against two leading MI attack strategies: the algorithms of Plug \& Play Attack~\cite{ppa22icml} (PPA) and training with entropy loss of KEDMI~\cite{kedmi21iccv} (HLoss).
Given that KEDMI's algorithms, aside from HLoss, are tailored specifically for DCGAN~\cite{ppa22icml}, and PPA is recognized as a versatile, open-source framework for incorporating StyleGAN in MI attacks~\cite{ppa22icml,mi23mm}, we adopt PPA's StyleGAN-based architecture to implement both HLoss and our proposed approach.
Note that while several MI studies~\cite{gmi20cvpr,mirror22ndss,vmi21nips,rethink23cvpr,mi23mm} focus on the latter phase of MI attacks (i.e., fine-tuning GAN latent codes), our approach primarily enhances the GAN transfer learning phase of MI Attacks.
Thus, our approach is compatible with these works, as well as future MI attack strategies centered on latent code optimization.

\noindent
\textbf{Competing Evaluation metrics.}\quad
We evaluate the attacks in terms of DDCS$_{\text{avg}}$ and DDCS$_{\text{best}}$ as introduced in Algorithm~\ref{alg:ddcs}, as well as other standard metrics including accuracy, feature distance, FID, and coverage from~\cite{densitycoverage,vmi21nips}.
Additionally, an Inception-v3~\cite{inceptionv3} model is trained on the target dataset as the evaluator on top-1 accuracy (Acc@1), top-5 accuracy (Acc@5) and feature distance (Dist).
A publicly available pre-trained Inception-v3 model is used to calculate the coverage and FID between the reconstructed samples and target samples.
Furthermore, to measure LPIPS distance, a default VGG network setting is adopted.

\subsection{DDCS Reveals Per-example Vulnerability Against MI Attacks}
\label{sec:exp:per_examples}

Previous works fail to pinpoint which sample from $\mathcal{D}_{tar}$ is the most similar to a given reconstructed sample.
As a result, in the visualization, one must manually select the most similar sample from $\mathcal{D}_{tar}$, which is not only best guaranteed but also hard to interpret.
On the other hand, \textbf{DDCS can indicate how vulnerable each sample is under the MI attack in an unsupervised manner}.
Specifically, recall that each sample from $\mathcal{D}_{tar}$ is assigned a reconstruction distance to indicate its degree of restoration by $\mathcal{D}_{rec}$ in either worst case or average case.
Therefore, we can easily build the reconstruction pairs between $\mathcal{D}_{tar}$ and $\mathcal{D}_{rec}$ by searching for samples in $\mathcal{D}_{rec}$ that can achieve the smallest reconstruction distance to a given sample in $\mathcal{D}_{tar}$.

\begin{figure}[htbp]%[htbp]
%\vspace{-0.2in}
	\centering
	\includegraphics[width=0.7\linewidth]{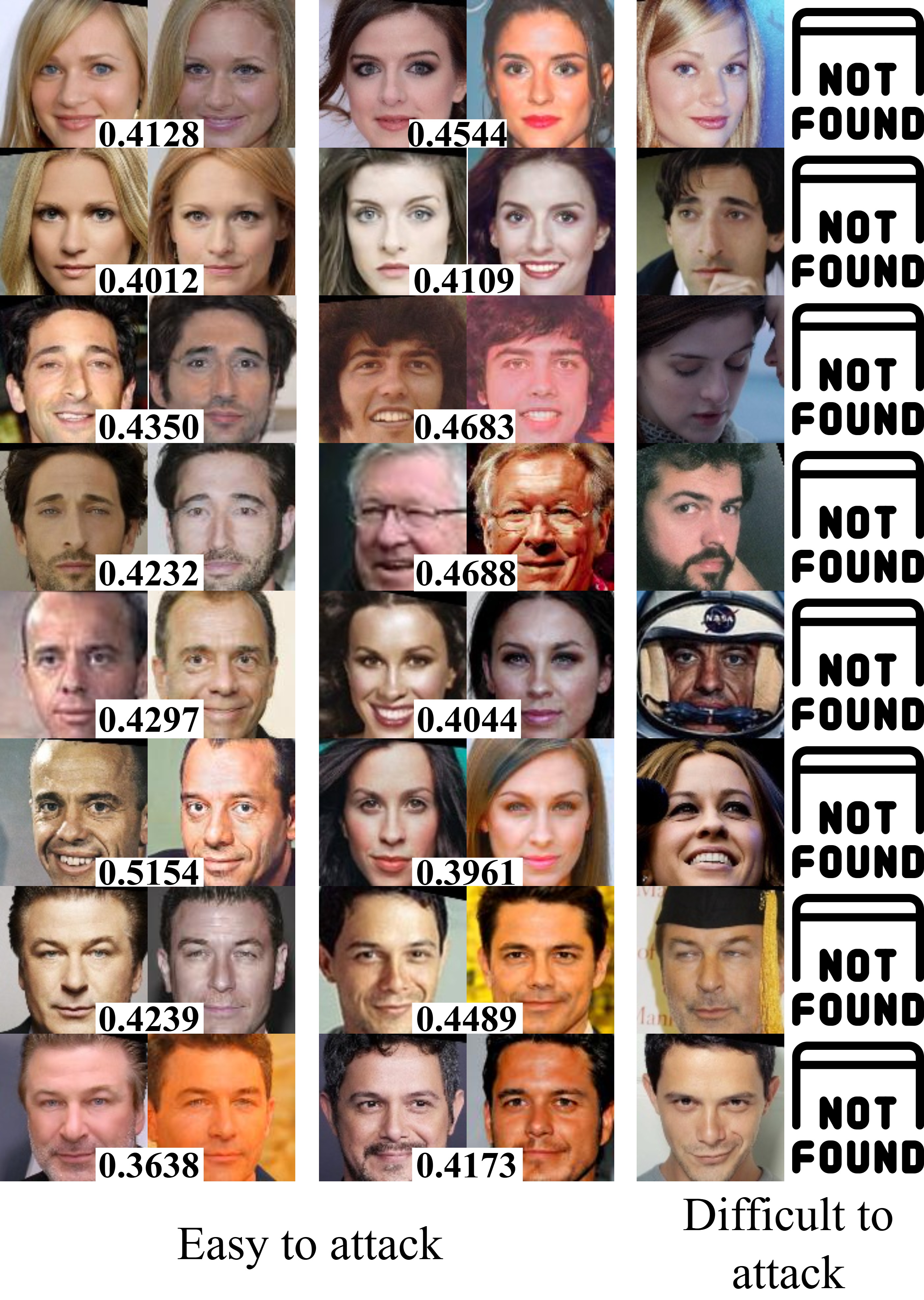}
 \vspace{-0.1in}
	\caption{Visualization of reconstruction pairs between target (left) and reconstructed (right) samples for VGG16BN-UMDFaces with reconstruction distance attached on their bottom right corner. Samples on the right most column have no reconstruction pairs.}
	\label{fig:eye_test}
 %\vspace{-0.2in}
\end{figure}

In Figure~\ref{fig:eye_test}, we present several samples from $\mathcal{D}_{tar}$ that are successfully matched by samples from $\mathcal{D}_{rec}$ together with their reconstruction distance. Samples that are not matched by any samples from $\mathcal{D}_{rec}$ are shown in the rightmost column.
Results are averaged from 10 independent runs to account for randomness.
These matched samples can be regarded as samples that are vulnerable to MI attacks and are more likely to leak their private information.
The data or model owner can thus focus on protecting these samples, by perturbing their values or removing them from the training.
Besides, we find that most samples in $\mathcal{D}_{tar}$ are unmatched, and these samples usually have more complex features than paired samples.
As such, future works could target at those difficult-to-attack samples to achieve a better coverage of MI attacks.

\begin{table*}[htb]
	\centering
    \resizebox{0.8\textwidth}{!}{
	\begin{tabular}{ | c | c | c c c c c | c c | }
		\hline
		Model & Attack & $\uparrow$ Acc@1 & $\uparrow$ Acc@5 & $\downarrow$ Dist & $\downarrow$ FID & $\uparrow$ Coverage & $\uparrow$ DDCS$_{\text{avg}}$ & $\uparrow$ DDCS$_{\text{best}}$\\
		\hline
		\multirow{3}{*}{VGG16BN} & HLoss & 49.64\% & 69.10\% & 15978.52 & 64.00 & 0.1601 & 0.4162 & 0.4308 \\
        & PPA & \bf 97.67\% & \bf 99.89\% & \bf 7044.45 & 48.35 & 0.3213 & 0.5046 & 0.5240 \\
		& Ours & \bf 97.67\% & 99.86\% & 7065.28 & \bf 47.64 & \bf 0.3907 & \bf 0.6772 & \bf 0.7114 \\
		\hline
		\multirow{3}{*}{ResNet50} & HLoss & 35.15\% & 57.01\% & 18086.90 & 51.48 & 0.1727 & 0.4118 & 0.4243 \\
        & PPA & \bf 81.77\% & 94.84\% & \bf 10605.46 & 47.10 & 0.3234 & 0.4901 & 0.5088 \\
		& Ours & 81.65\% & \bf 95.01\% & 10642.02 & \bf 45.83 & \bf 0.3951 & \bf 0.6632 & \bf 0.6961 \\
		\hline
		\multirow{3}{*}{IR50-SE} & HLoss & 31.58\% & 49.47\% & 18323.38 & 79.28 & 0.1303 & 0.4238 & 0.4364 \\
        & PPA & 85.65\% & 95.64\% & 9728.30 & 47.55 & 0.3186 & 0.5146 & 0.5339 \\
		& Ours & \bf 86.12\% & \bf 95.92\% & \bf 9668.79 & \bf 46.18 & \bf 0.3901 & \bf 0.7017 & \bf 0.7367 \\
		\hline
		\multirow{3}{*}{AlexNet} & HLoss & 16.68\% & 36.07\% & 19990.63 & 62.06 & 0.1543 & 0.3901 & 0.4031 \\
        & PPA & 53.96\% & \bf 78.54\% & \bf 13710.45 & 45.62 & 0.2979 & 0.4600 & 0.4762 \\
		& Ours & \bf 53.97\% & 78.43\% & 13751.00 & \bf 44.11 & \bf 0.3696 & \bf 0.6193 & \bf 0.6481 \\
		%\hline
		%\multirow{3}{*}{ResNet50} & PPA & - & - & - & - & - & - & - & - \\
		%& HLoss & - & - & - & - & - & - & - & - \\
		%& Ours & - & - & - & - & - & - & - & - \\
		\hline
	\end{tabular}
    }
	\caption{Comparison on UMDFaces dataset between our approach (Ours), HLoss and PPA accross different metrics. $\uparrow$ and $\downarrow$ mean the higher the better and the lower the better, respectively. The best values for each metric and each model are in bold.}
	\label{exp:result:umdfaces}
\end{table*}

\subsection{Comparison of Various Evaluation Metrics}
In this subsection, we experimentally study the evaluation and robustness of DDCS against limitations discussed previously.
Specifically, we build a customized $\mathcal{D}_{rec}$ to simulate an MI attack that are very close to the ideal MI attack and can reconstruct $\mathcal{D}_{tar}$ very well.
As such, we can then observe how DDCS successfully evaluates this customized attack and how existing metrics fail to achieve it.
Two customized datasets, denoted as D1 and D2, are constructed based on UMDFaces and the settings as follows:

\noindent
D1. \quad
This dataset represents a $\mathcal{D}_{rec}$ of perfect distance to $\mathcal{D}_{tar}$ but poor diversity. To construct D1, we sub-sample a fixed number of images for every label, so that those samples have their identical matches in original $\mathcal{D}_{tar}$ and the diversity of D1 can be manually controlled.

\noindent
D2.\quad
This dataset represents a $\mathcal{D}_{rec}$ of perfect distance and diversity to $\mathcal{D}_{tar}$ but with a modified sample distribution different from $\mathcal{D}_{tar}$. To construct D2, we create a fixed number of redundant samples from $\mathcal{D}_{tar}$ to result in different distribution between $\mathcal{D}_{tar}$ and $\mathcal{D}_{rec}$.

In D1, we simulate the condition of varied diversity in $\mathcal{D}_{rec}$, and thus DDCS is compared with accuracy in this case.
As for D2, we test the metrics' robustness against distribution change of $\mathcal{D}_{rec}$, a common phenomenon in MI attacks, where we choose FID as the competing metric for DDCS.
Since both customized datasets can achieve perfect distance to $\mathcal{D}_{tar}$, we regularize the range of DDCS by setting $c$ in Algorithm~\ref{alg:ddcs} to 1.0.

\begin{figure}[htb]%[htbp]
	\centering
	\includegraphics[width=0.7\linewidth]{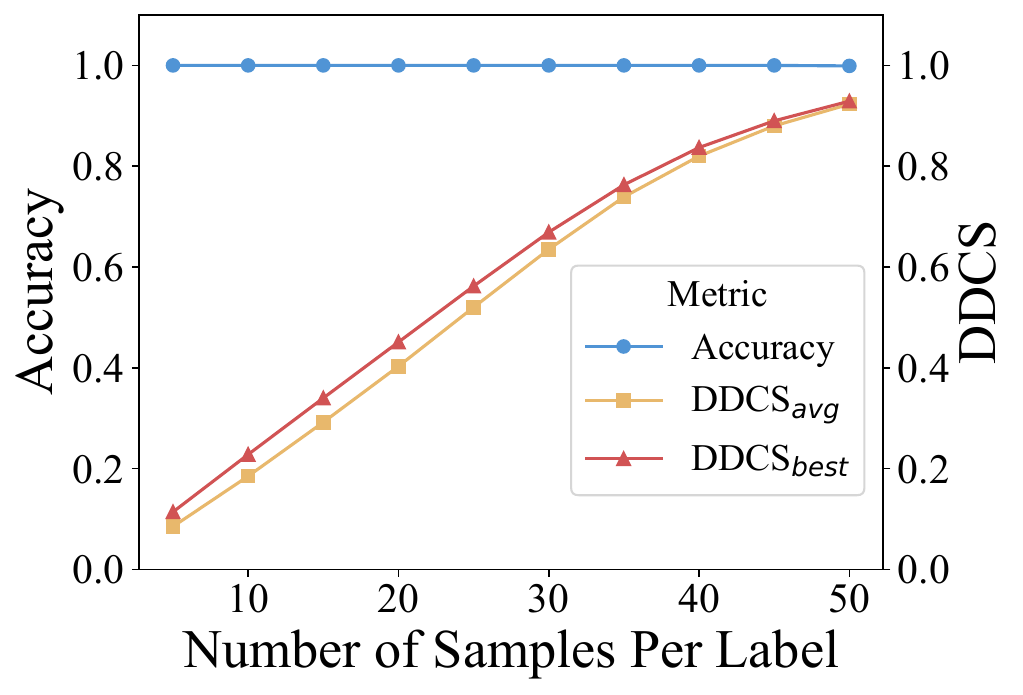}
 \vspace{-0.1in}
	\caption{Evaluation results for various metrics on a customized UMDFaces dataset, in which the diversity is controlled with different number of samples per label. }
	\label{fig:rich_poor}
 \vspace{-0.2in}
\end{figure}

\begin{figure}[htb]%[htbp]
	\centering
	\includegraphics[width=0.7\linewidth]{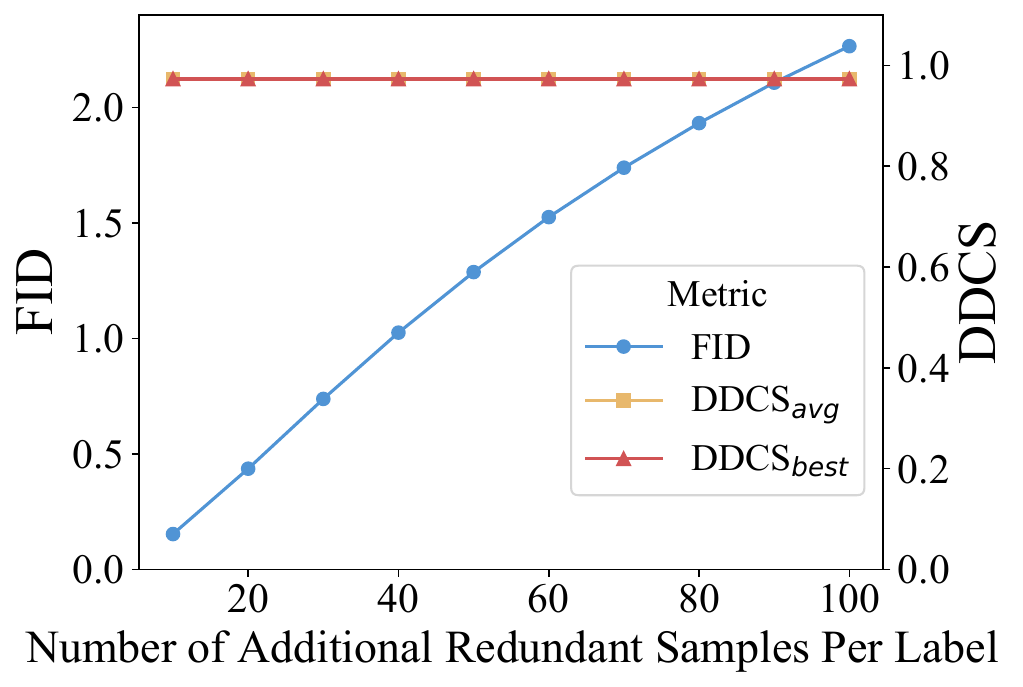}
 \vspace{-0.1in}
	\caption{Evaluation results for various metrics on a customized UMDFaces dataset, in which the distribution is controlled with different number of redundant samples per label. }
	\label{fig:eval_redundant}
 %\vspace{-0.2in}
\end{figure}

Figure~\ref{fig:rich_poor} and~\ref{fig:eval_redundant} are the results for D1 and D2 respectively.
In Figure~\ref{fig:rich_poor}, as more images are sampled, DDCS successfully captures the increasing diversity of $\mathcal{D}_{rec}$ and thus increases. On the other hand, since $\mathcal{D}_{rec}$ has a very small distance from the samples of $\mathcal{D}_{tar}$, the accuracy approaches 100\% but it ignores the diversity of $\mathcal{D}_{rec}$.
In Figure~\ref{fig:eval_redundant}, since FID is sensitive to the changes of sample distribution, it always grows with the addition of additional redundant samples. Thanks to the $\mathcal{D}_{rec}$-oriented evaluation, DDCS is robust against the varied distribution and stays in a high and stable range.

\subsection{Comparison Results of MI Attacks}

Table~\ref{exp:result:umdfaces} shows the comparison results on UMDFaces between Ours, HLoss and PPA. %Table~\ref{exp:result:stanforddogs} in the Appendix shows the result of Stanford Dogs.
We keep two decimal places for numbers greater than 10, and four decimal places for the rest. %Their detailed explanations are attached in the Appendix.

\subsection{Per-label Reconstruction of MI Attacks}
\begin{figure}[htb]%[htbp]
%\vspace{-0.1in}
	\centering
	\includegraphics[width=0.9\linewidth]{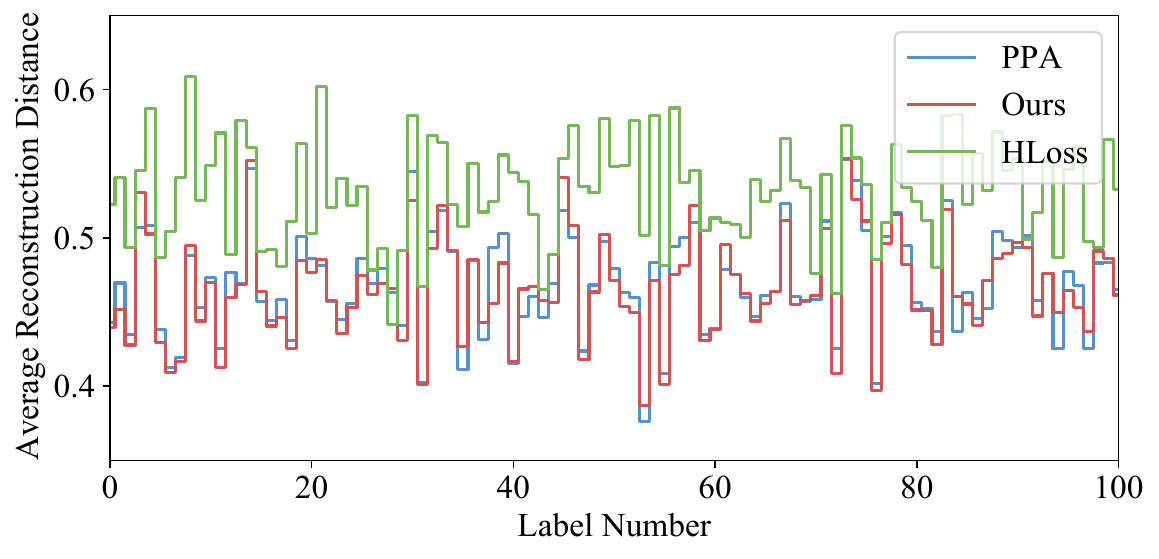}
 %\vspace{-0.1in}
	\caption{Average reconstruction distance for matched samples in each label with VGG16BN-UMDFaces and three attacks (PPA, HLoss and Ours).}
	\label{fig:ddcs_distance}
 \vspace{-0.1in}
\end{figure}

In this subsection, we study the reconstruction degree of MI attacks in a label-wise manner through visualizing the reconstruction distance from DDCS.
Among three attacks discussed in UMDFaces dataset and VGG16BN model scenario, Figure~\ref{fig:ddcs_coverage} plots the proportion of target samples that have at least one matching to $\mathcal{D}_{rec}$, and Figure~\ref{fig:ddcs_distance} illustrates the per-label reconstruction distance by averaging the cumulative distance of these matched target samples.
We observe that our method significantly enhances the proportion of those samples that are reconstructed by $\mathcal{D}_{rec}$, while achieving a similar reconstruction distance to PPA, thereby improving DDCS.
On the other hand, suffering from entropy loss with soft-constraint that deteriorates both the image quality and generative power, HLoss achieves a much higher average reconstruction distance and thus a lower DDCS, though it can cover some labels in Figure~\ref{fig:ddcs_coverage}.

%As the reconstruction distance of a sample decreases, there is an increasing risk of leakage of the privacy information it contains.

%\subsection{Improvements on DDCS for Previous Techniques}
%Another interesting question is how the previous attacks perform on DDCS?
%To begin with, starting from GMI, there have been several significant improvements in MI attacks, including the use of poincare distance, initial selection and final selection.
%We measure the improvement of these methods on DDCS$_\text{avg}$ and DDCS$_\text{best}$ and present the result in Table.
%Since DDCS comprehensively measures multiple characteristics of MI attacks, these techniques also achieves descent improvements DDCS.

%\subsection{Limitation of A Single GAN}
%To further illustrate the limitations of a single GAN for MI attacks, we study whether similar performance to GAN Augmentation can be achieved by increasing the number of generated samples for baseline.

%\subsection{More Augmented GANs}

%\subsection{HVP of Image Manifold}

%\subsection{Indenpence of $\lambda_h$} 
\section{Conclusion}
\label{sec:conclusion}
In this paper, we point out several limitations in the existing evaluation metrics of MI attacks and propose a novel metric, Diversity and Distance Composite Score (DDCS), to alleviate those limitations and encourage a more comprehensive evaluation of MI attacks.
%Focusing on sample-level privacy instead of label level privacy, DDCS accounts for the reconstruction distance of each target sample so as to consider multiple important characteristics of MI attacks and be robust on redundant samples from the reconstructed dataset.
%It is worth noting that DDCS assists a better analysis of sample vulnerability against MI attacks and indicates the low coverage in recent MI attacks.
Furthermore, to enhance existing MI attacks, we further propose a GAN augmentation framework with transfer learning for state-of-the-art MI attacks.
%In GAN augmentation, a pre-trained GAN with natural gradient on entropy loss is utilized to improve the generative ability of the MI attack.
%Extensive experiments on face identification and dog breed classification confirm the comprehensiveness and robustness of DDCS, as well as the effectiveness of our approach.
Overall, we emphasize the importance of informing the academic community about potential threat models of MI attacks and introducing new perspectives on privacy measurement, to foster the development of more robust privacy-preserving ML algorithms.

%Applying other techniques for improving GAN transfer learning~\cite{freezed20,lecamdiv21cvpr} to MI attacks are interesting for further discussion. 

%\newpage
\section{Acknowledgments}
This work was supported by the National Natural Science Foundation of China (Grant No: 92270123, 62072390, and 62372122), and the Research Grants Council, Hong Kong SAR, China (Grant No: 15203120, 15226221, 15209922, 15210023, and C2004-21GF). We appreciate Mr Andy Cheung Yat-ming for constructive discussions in the natural gradient descent part and anonymous reviewers for their constructive comments.

\bibliography{aaai25}
%\newpage
\appendix
%\clearpage
%\setcounter{page}{1}
%\maketitlesupplementary

% Classification accuracies of victim models

\section*{Appendix}

\section{Related Work}
\label{related}
\textbf{MI attacks with DCGAN.}\quad MI attacks are broadly classified into black-box and white-box categories based on the attacker's access to the target neural network's parameters.
Initial white-box MI attacks~\cite{mi15} lacked stringent optimization constraints, leading to the generation of blurred and unrealistic samples.
To enhance sample realism, \cite{gmi20cvpr} integrated DCGAN into the MI attack framework by optimizing the latent codes of generated samples.
The use of explainable artificial intelligence tools in MI attacks was explored by \cite{xai21iccv}, while KEDMI~\cite{kedmi21iccv} advanced the field by developing an inversion-specific DCGAN featuring a multi-task discriminator and entropy loss.
Further study~\cite{rethink23cvpr} suggests improved optimization objective and model augmentation to mitigate the sub-optimum and overfitting problem.

\noindent
\textbf{MI attacks with StyleGAN.}\quad Our work primarily investigates MI inversions utilizing StyleGAN~\cite{stylegan2_20cvpr}, recognized for setting the benchmark in generating image priors for MI attacks.
\cite{vmi21nips} approached MI attacks through a probabilistic lens, employing StyleGAN to elevate image quality.
The study by \cite{mirror22ndss} introduced heuristic improvements for StyleGAN-based MI attacks in both white-box and black-box scenarios.
To overcome challenges like vanishing gradients and distributional shifts, the Plug \& Play Attack~\cite{ppa22icml} was proposed, incorporating Poincar\'e loss and a transformation-based selection mechanism.
Enhancements to this method were later made by \cite{mi23mm}, leveraging historically learned knowledge.

\noindent
\textbf{Black-box MI attacks.}\quad In the context of black-box MI attacks, where the attacker lacks access to the target model's parameters, \cite{blackbox19ccs} developed a surrogate neural network to invert the undisclosed victim model. The introduction of boundary repulsion by \cite{blackbox22cvpr} enabled MI attacks to function using only the label outputs of the victim model. Further strengthening of black-box MI attacks was achieved by \cite{blackbox23tdsc} through the integration of semantic loss regularization and adversarial example injection. The study by \cite{mi23iccvw} addressed the challenge of attacking black-box face recognition models by harnessing the stochastic properties of the denoising diffusion process from diffusion models.

\noindent
\textbf{Gradient inversion attacks.}\quad Gradient Inversion (GI)~\cite{wr1} is fundamentally different from MI attacks, where GI is concentrated on those scenarios in federated learning requiring gradient information. Notably, recent works on GI~\cite{wr2, wr3, wr4, wr6} still inherit the same evaluation metrics from previous MI works, which makes our work unique. Specifically, \cite{wr1} relaxes the strong assumption about the GI attacks and proposes a more practical scenario. Other works~\cite{wr2, wr3, wr4, wr6} propose GI attacks in FL using DCGAN and BigGAN and are assessed with basic accuracy and distance metrics as the evaluation metrics.

\section{Differences between DDCS and KNN-dist}
One might question the differences between DDCS and K Nearest Neighbor distance (KNN-dist) since they both find closet pairwise match in a sample-level manner.
However, KNN-dist differs significantly from DDCS in evaluative approach. KNN-dist finds the closest match in target dataset for each reconstructed sample, while DDCS has two unique advantages. First, DDCS allows certain samples in target dataset (i.e., hard-to-attack samples shown in Figure~\ref{fig:eye_test}) to remain unmatched (Line~\ref{alg:rec_dist} in Algorithm~\ref{alg:ddcs}), providing a more accurate privacy leakage assessment by recognizing sample-specific challenges~\cite{mif22sp}. Second, KNN-dist averages all reconstructed samples, ignoring sample diversity, which may incentivize attacks to generate a single high-quality sample. In contrast, DDCS averages over all target samples, thus encouraging MI attacks to improve both diversity and accuracy for higher coverage. 

\section{NGD-based Approach}
Denote the Euclidean gradient for the entropy loss $\mathcal{L}_{\text{adv}}$ regarding the parameters of generator $G$ as $\frac{\partial \mathcal{L}_{\text{adv}}}{\partial G}$ and the projected gradient as $P(\cdot)$.
Since the enropty loss is obtained by feeding the generated images $G(z)$ to the victim model with the latent codes $z$, we can expand $\frac{\partial \mathcal{L}_{\text{adv}}}{\partial G}$ with the chain rule:
\begin{equation}
	\frac{\partial \mathcal{L}_{\text{adv}}}{\partial G} = \frac{\partial \mathcal{L}_{\text{adv}}}{\partial G(z)} \cdot
	\frac{\partial G(z)}{\partial G}
	\label{eq:chain_rule}
\end{equation}

$\frac{\partial \mathcal{L}_{\text{adv}}}{\partial G(z)}$ denotes the gradient of $\mathcal{L}_{\text{adv}}$ w.r.t. the generated images $G(z)$, which will be ill-conditioned under the soft constraint during the optimization.
Such unconstrained update direction will bring artifacts to $G(z)$ and thus causes them to deviate from the original manifold.

Recall the natual gradient~\cite{ngd} of $\frac{\partial G(z)}{\partial G}$, represented as $\Delta G(z)$, is computed as:
\begin{equation}
	\Delta G(z) = F^{-1} \frac{\partial \mathcal{L}_{\text{adv}}}{\partial G(z)},
	\label{eq:ngd_def}
\end{equation}
where $F^{-1}$ denotes the inverse of FIM, which is approximated as the inverse of metric tensor~\cite{rogan18cvpr}:
\begin{equation}
	F^{-1} \approx H^{-1}_{G(z)}(d^2(G(z),G(z_0)))
	\label{eq:ngd_approx} = v\lambda^{-1}v^T,
\end{equation}
where $G(z_0)$ is another batch of generated images with latent codes $z_0$ that are close to $z$. $H_{G(z)}(d^2(G(z),G(z_0)))$ is computed by taking the second order partial derivative of squared LPIPS distance $d^2(G(z),G(z_0))$ w.r.t. $G(z)$. The Hessian can then be decomposed as $v\lambda v^T$, using the Hessian's eigenvalue $\lambda$ and eigenvector $v$.

Summarizing Equations~\ref{eq:chain_rule},~\ref{eq:ngd_def} and~\ref{eq:ngd_approx}, the projected gradient of $\mathcal{L}_{\text{adv}}$ can be expressed as:
\begin{equation}
\begin{split}
	P(\frac{\partial \mathcal{L}_{\text{adv}}}{\partial G})=F^{-1} \frac{\partial \mathcal{L}_{\text{adv}}}{\partial G(z)} \cdot \frac{\partial G(z)}{\partial G} \\
	\approx H^{-1}_{G(z)}(d^2(G(z),G(z_0))) \frac{\partial \mathcal{L}_{\text{adv}}}{\partial G(z)} \cdot \frac{\partial G(z)}{\partial G}
\end{split}
\end{equation}

\section{GAN Augmentation Algorithm}
\begin{figure}[htb]
	\centering
	\includegraphics[width=1.0\linewidth]{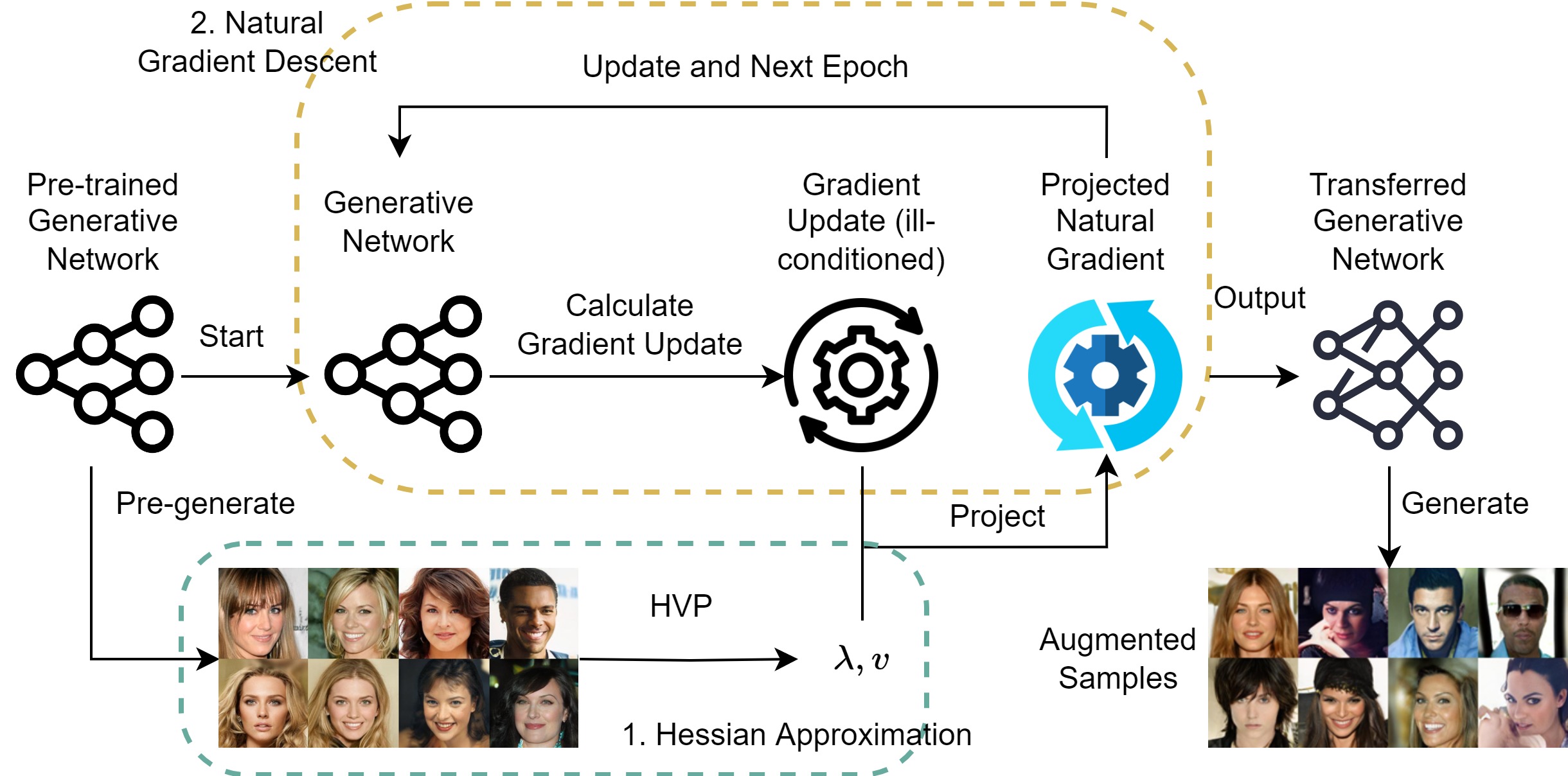}
	\caption{Overview of our transfer learning attack framework.}
	\label{fig:overview_gan_augmentation}
\end{figure}

Our framework leverages DDCS's insights, showing existing MI attacks’ low coverage of the target dataset. This inspired expanding sample diversity by utilizing the victim model in fine-tuning the generator, forming the basis of our transfer framework.
Figure~\ref{fig:overview_gan_augmentation} illustrates the overview of our framework. In our attack, we assume the attacker has access to some public AI models (\emph{e.g.}, from open source or model marketplace) so that he/she could download the pre-trained generative network. Given a victim model with its parameters, our framework will first perform Hessian Vector Product (HVP) to obtain an approximated Hessian. In each epoch, the approximated Hessian will help to project ill-conditioned gradient into its natural gradient. As a result, the fine-tuned StyleGAN can generate more diverse samples related to the private information of the victim model while preserving the image quality. Such approach increases per-epoch runtime from 285 to 829 seconds, a manageable overhead with fewer than 20 epochs.

Algorithm~\ref{alg:gan_aug} illustrates the complete process of our GAN augmentation approach, where the subscript $_{\text{van}}$ denotes the vanilla transfer and $_{\text{adv}}$ denotes the transfer with entropy loss.
For every step, the generator $G$ will produce a batch of fake images $G(z)$ (Line~\ref{alg2:batch_fake}) to perform a normal training (Line~\ref{alg2:normal_train}) and obtain the entropy loss (Line~\ref{alg2:entropy}).
We update $G$ with the projected gradient of $\mathcal{L}_{\text{adv}}$ using Equation~\ref{eq:ngd} (Line~\ref{alg2:ngd}).

\begin{algorithm}[htb]
	\caption{GAN Augmentation}
	\label{alg:gan_aug}
	\textbf{Input}: Pre-trained generator $G$ and discriminator $D$, auxiliary dataset $\mathcal{D}_{aux}$, victim model $V$ \\
	\textbf{Output}: Recovered dataset $\mathcal{D}_{rec}$
	\begin{algorithmic}[1]
	\STATE Generate $\mathcal{D}_{\text{vanilla}}$ with $G$ and $V$
	
	\FOR {each iteration step}
    \STATE Generate a batch of images $G(z)$ with random latent code $z$ \label{alg2:batch_fake}
	\STATE Update $G$ and $D$ with $\mathcal{D}_{aux}$ and $G(z)$ \label{alg2:normal_train}
	\STATE Calculate adversarial loss $\mathcal{L}_{\text{adv}}$ with $V$ and $G(z)$ \label{alg2:entropy}
	\STATE Update $G$ with gradient $H^{-1}_{G(z)}(d^2(G(z),G(z_0))) \frac{\partial \mathcal{L}_{\text{adv}}}{\partial G(z)} \cdot \frac{\partial G(z)}{\partial G}$ \label{alg2:ngd}
	\ENDFOR
	
	\STATE Generate $\mathcal{D}_{\text{adv}}$ with $G$ and $V$
	
	\STATE $\mathcal{D}_{rec} \gets \{\mathcal{D}_{\text{vanilla}}, \mathcal{D}_{\text{adv}}\}$
 \end{algorithmic}
\end{algorithm}

%[TODO: I think a summary of your total augmenting mechanism using pseudo-code will be nice here to conclude this section.]

\section{Improvements of DDCS by Producing More $\mathcal{D}_{rec}$}
\begin{figure}[htb]%[htbp]
	\centering
	\begin{minipage}[t]{0.495\linewidth}
		\includegraphics[width=\linewidth]{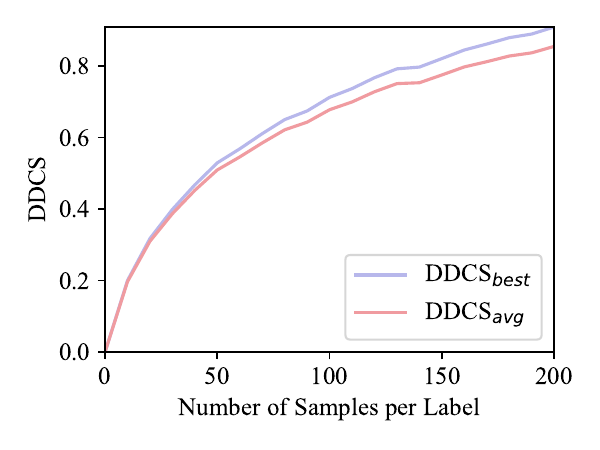}
		%\captionsetup{font={scriptsize}}
		%\caption*{(a)}
	\end{minipage}
	\begin{minipage}[t]{0.495\linewidth}
		\includegraphics[width=\linewidth]{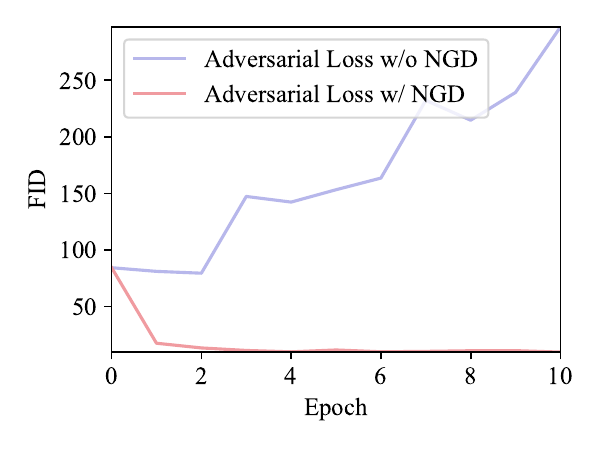}
		%\captionsetup{font={scriptsize}}
		%\caption*{(b)}
	\end{minipage}
	\caption{(a). left sub-figure indicates the corresponding DDCS$_\text{avg}$ and DDCS$_\text{best}$ when the attacker uses a different number of generated samples for each label. (b). right sub-figure shows the change of FID during the training of StyleGAN with different training settings. We use the VGG16BN-UMDFaces setting for both figures.}
	\label{exp:ms_iq}
\end{figure}

Intuitively, generating more diverse samples by a generative network can result in more samples from $\mathcal{D}_{tar}$ being reconstructed by $\mathcal{D}_{rec}$.
As shown in Figure~\ref{exp:ms_iq} (a), as the number of samples generated by the attacker on each label increases, both DDCS$_{\text{avg}}$ and DDCS$_{\text{best}}$ demonstrate a significant improvement at the beginning and the improvement tends to decelerate when the number of samples are great enough.
This indicates that DDCS will increase with larger sample size per label of $\mathcal{D}_{rec}$, especially when the sample size is originally small.
Nevertheless, such relationship is not linear, which in turn suggests that more redundant samples are also produced. As such, to further improve the strength of MI attacks, we should turn to other approaches such as GAN augmentation.

\section{Image Quality of Training with Adversarial Loss}
In Figure~\ref{exp:ms_iq} (b), we measure FID, which in this section represents the naturalness of generated images by comparing the generated images with real images. The red dash line and the black solid line denote entropy loss with and without our training approach in Section~\ref{sec:gan_aug:ngd}, respectively. Although there is a noticeable improvement of entropy loss on DCGAN~\cite{kedmi21iccv,rethink23cvpr}, the entropy loss continuously damages the naturalness of images on deeper generative networks, \emph{i.e.}, StyleGAN.
On the other hand, our approach preserves the images into the manifold and ensures the convergence of StyleGAN when training with adversarial loss.
Figure~\ref{fig:transfer_quality} illustrates the visual effect, in which our approach removes most visible artifacts caused by the entropy loss.
%[TODO: any observation here?]

\section{Results for Stanford Dogs}
\begin{table*}[htb]
	\centering
	\begin{tabular}{ | c | c | c c c c c | c c | }
		\hline
		Model & Attack & $\uparrow$ Acc@1 & $\uparrow$ Acc@5 & $\downarrow$ Dist & $\downarrow$ FID & $\uparrow$ Coverage & $\uparrow$ DDCS$_{\text{avg}}$ & $\uparrow$ DDCS$_{\text{best}}$ \\
		\hline
		\multirow{3}{*}{VGG16BN} & HLoss & 17.93\% & 42.32\% & 11984.44 & 152.08 & 0.0033 & 0.1434 & 0.1450 \\
        & PPA & \bf 45.70\% & \bf 77.01\% & \bf 8616.43 & 57.63 & 0.0569 & 0.1777 & 0.1797 \\
		& Ours & 44.23\% & 76.00\% & 8766.75 & \bf 57.21 & \bf 0.0782 & \bf 0.2787 & \bf 0.2830 \\
		\hline
		\multirow{3}{*}{ResNet50} & HLoss & 13.67\% & 41.80\% & 11456.74 & 177.82 & 0.0014 & 0.1512 & 0.1530 \\
        & PPA & \bf 52.64\% & \bf 82.00\% & \bf 8139.98 & 53.67 & 0.0650 & 0.1663 & 0.1682 \\
		& Ours & 52.17\% & 81.90\% & 8194.13 & \bf 51.94 & \bf 0.0930 & \bf 0.2604 & \bf 0.2644 \\
		\hline
		\multirow{3}{*}{IR50-SE} & HLoss & 9.37\% & 23.22\% & 12325.54 & 199.59 & 0.0006 & 0.1339 & 0.1354 \\
        & PPA & \bf 38.27\% & \bf 71.73\% & \bf 8862.13 & 58.10 & 0.0499 & 0.1585 & 0.1603 \\
		& Ours & 37.88\% & 71.11\% & 8907.61 & \bf 55.46 & \bf 0.0741 & \bf 0.2537 & \bf 0.2573 \\
		\hline
		\multirow{3}{*}{AlexNet} & HLoss & 13.80\% & 35.80\% & 11655.26 & 135.11 & 0.0047 & 0.1464 & 0.1480 \\
        & PPA & 28.12\% & 54.25\% & 9950.77 & 60.80 & 0.0389 & 0.1613 & 0.1632 \\
		& Ours & \bf 28.17\% & \bf 54.93\% & \bf 9938.49 & \bf 57.17 & \bf 0.0601 & \bf 0.2513 & \bf 0.2552 \\
		\hline
	\end{tabular}
	\caption{Comparison on Stanford Dogs dataset between our approach (Ours), HLoss and PPA accross different metrics. $\uparrow$ and $\downarrow$ denote the higher the better and the lower the better, respectively. The best values for each metric and each model are in bold.}
	\label{exp:result:stanforddogs}
 %\vspace{-0.2in}
\end{table*}

\begin{figure}[htbp]%[htbp]
	\centering
	\includegraphics[width=0.95\linewidth]{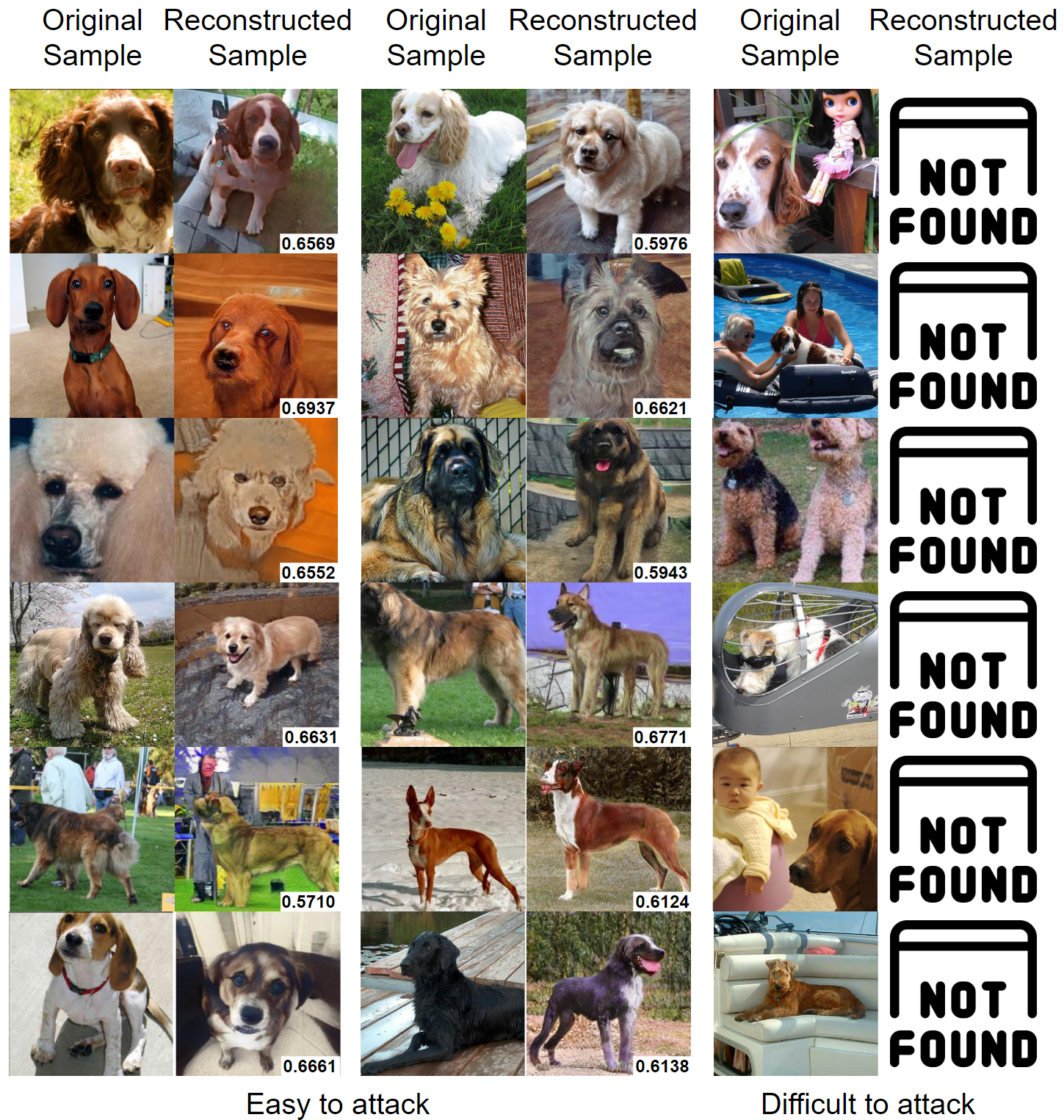}
	\caption{Visualization of reconstruction pairs between target and reconstructed samples for ResNet50-StanfordDogs with reconstruction distance attached on their bottom right corner. Samples on the right most column have no reconstruction pairs.}
	\label{fig:eye_test_dogs}
\end{figure}

Figure~\ref{fig:eye_test_dogs} represents several target samples and their reconstruction pairs.
Since dog breed classification has a lower MI attack success rate than face recognition in terms of accuracy, FID and coverage, the reconstruction distances between $\mathcal{D}_{tar}$ and $\mathcal{D}_{rec}$ are also larger than that in Figure~\ref{fig:eye_test}.
However, many important private details are still captured by the attack, such as the dog's posture and the general scene it is in.
Future work could focus on mining the privacy of tasks beyond face recognition including dog breed classification, and reconstructing samples with more complex features as shown in the right most column in Figures~\ref{fig:eye_test} and~\ref{fig:eye_test_dogs}.

\section{Implementation Details}
In selecting the distance function $d(\cdot,\cdot)$ for DDCS in Algorithm~\ref{alg:ddcs}, we adopt Learned Perceptual Image Patch Similarity (LPIPS)~\cite{lpips18}. Other distance functions (\emph{e.g.}, KNN distance~\cite{kedmi21iccv}) assess samples from a distributional perspective. In contrast, LPIPS judges the semantic differences between individual samples, aligning with our focus on sample-level privacy.
Additionally, to mitigate the dependency on pre-trained networks, DDCS can incorporate a simplified version of the LPIPS metric~\cite{lpips21}.
Since we target at a general ML scenario where sample-level privacy of $\mathcal{D}_{tar}$ are much more important than label-level privacy, we allow label intersection between the training datasets and the auxiliary datasets (\emph{e.g.}, the label of \textit{Airedale} might be appeared in both target and auxiliary datasets). As for training framework, we adopt Face.evoLVe~\cite{faceevolve21} to train our target models, and fork the PyTorch training framework~\footnote{https://github.com/rosinality/stylegan2-pytorch} of StyleGAN with Adaptive Discriminator Augmentation~\cite{stylegan2ada_20nips} with default settings.

Regarding the dataset selection, we primarily focus on face datasets in this paper, as they contain numerous sensitive attributes and effectively illustrate the risks associated with MI attacks. Datasets like the disease or dog breed dataset are less suited to sample-level privacy demonstration, as individual images contain limited private information. Additionally, most existing pre-trained StyleGAN models are trained on face datasets. Since we make minimal assumptions about the pre-trained generator, we use the default settings from the official StyleGAN repository~\footnote{https://github.com/NVlabs/stylegan2} and fine-tune the model with human face and dog breed datasets.
Furthermore, StyleGAN is pre-trained on Flickr-Faces-HQ, whereas we utilize UMDFaces for training the victim model to minimize sample overlap between the target and pre-training datasets. UMDFaces captures varied real-world conditions~\cite{umdfaces}, while Flickr-Faces-HQ spans diverse attributes like age, ethnicity, and facial features~\cite{stylegan2ada_20nips}. Given these differences, achieving meaningful sample overlap is challenging. Even if overlap occurs, the victim model must first memorize these samples, as MI attacks focus on maximizing confidence in the victim model. Consequently, any sample memorized solely by StyleGAN, without memorization by the victim model, is unlikely.

\subsection{Comparison Results of MI Attacks}
These results in Tables~\ref{exp:result:umdfaces} and~\ref{exp:result:stanforddogs} suggest that our proposed method is more effective in reconstructing images and inverting the target models compared to other baseline MI attacks.
In the previous metrics, our attack achieve significant improvements on coverage and FID. These improvements are accounted by DDCS as well, so that our attack also improve DDCS very effectively.
Furthermore, DDCS of HLoss decreases significantly as all previous metrics are deteriorated in such attack, indicating the improvement of our approach on removing the artifacts introduced by HLoss in StyleGAN situation.
Notably, the results reflect that DDCS manages to evaluate multiple characteristics comprehensively to identify good MI attacks.
For example, in the case of VGG16BN-UMDFaces, our approach improves the coverage and FID to 0.3907 and 47.64 respectively, while maintaining the top-1 accuracy, top-5 accuracy and feature distance to 96.67\%, 99.86\% and 7065.28. As a result, DDCS$_{\text{avg}}$ and DDCS$_{\text{best}}$ of our approach are improved to 0.6772 and 0.7114.
In the same condition, PPA achieves inferior FID and coverage of 48.35 and 0.3213, consequently with lower DDCS$_{\text{avg}}$ and DDCS$_{\text{best}}$ of 0.4162 and 0.4308.
Despite the notable performance of the baseline attacks in previous metrics, their DDCS values have considerable room for improvement.
Recall that, compared to previous metrics, DDCS provides a more comprehensive evaluation for assessing MI attacks from the perspective of an ideal attack model, the lower performance of prior MI attacks on DDCS reflects their inherent limitations in targeting sample-level privacy. Thus, from the experimental results, we could observe that previous attacks underperform due to fundamental weaknesses not captured by existing metrics. Additionally, given DDCS's broader evaluation capacity, we encourage future MI attacks and defenses to strive for improved performance under DDCS to more accurately assess privacy risks in machine learning.

In top-1 accuracy, top-5 accuracy and feature distance, though these metrics have been shown defective on evaluating MI attacks in our paper, our method still achieves comparative or equivalent accuracy to the baseline models.
Unlike previous MI works on improving image naturalness, our framework enhances reconstruction coverage and preserves image quality, thus Acc and Dist fall short of quantifying the private samples our approach recovers. However, DDCS, FID and coverage successfully capture our framework's improvements, reflecting a meaningful advancement in MI attacks beyond traditional metrics.
These observations suggest that our approach mitigates the drawbacks from HLoss, so that the attack does not compromise the accuracy and feature distance of reconstructed samples.
As for distance metric on distributional similarity (\emph{i.e.}, FID and coverage), our method consistently achieves a lower FID than the PPA and HLoss, demonstrating higher distributional similarity between the recovered dataset and the target dataset.
Since both our method and baselines generate many redundant samples, we suggest one can implement supervised removal of redundant samples to reduce FID in the future.
Besides, the proposed method consistently achieves higher coverage values compared to the baselines, signifying better reconstruction on the target dataset and more realistic generative samples.

%Apart from the main comparisons, some side observations can be inferred from the tables.
%First, our method preserves a high consistency across different models and datasets.
%For all models (VGG16BN, ResNet50, IR50-SE and AlexNet) and all datasets, our method achieves similar improvement over the baseline in terms of most metrics, suggesting its robustness across various architectures and application scenarios.

Furthermore, we notice that in the results of Stanford Dogs, the overall performance of all attacks is lower than that in UMDFaces.
This is because compared to UMDFaces where faces are all well aligned and cropped, the feature distribution of Stanford Dogs dataset is much more intricate (\emph{e.g.}, dogs in the pictures are in different poses, different positions, and on different backgrounds).
We also observe that it is more difficult to attack AlexNet victim model than that of other architectures.
This is because compared to other more advanced architectures, AlexNet does not generalize well to the testing dataset and might remember some unnatural and trivial pixel-wise features from the training dataset.
Since the memorization of these victim models is poor, it in turn impacts on the success of MI attacks, which coincides with the findings of recent works~\cite{mldoctor22usenix,gmi20cvpr} that difficult training tasks can undermine the performance of MI attacks.

%We also find that the deterioration of adversarial loss without natural gradient is worse on complex tasks than simple tasks (\emph{i.e.}, dog breed classification on Stanford Dogs).
%Compared with UMDFaces, HLoss in Stanford Dogs is even more damaging to MI attacks, which can be reflected in FID, density and coverage of Table~\ref{exp:result:stanforddogs}.
%However, our method is still able to achieve a decent improvement in such cases, thanks to the protection of image quality by NGD.

\section{Redundant Samples from MI Attacks}
\begin{figure}[t]%[htbp]
	\centering
	\includegraphics[width=0.8\linewidth]{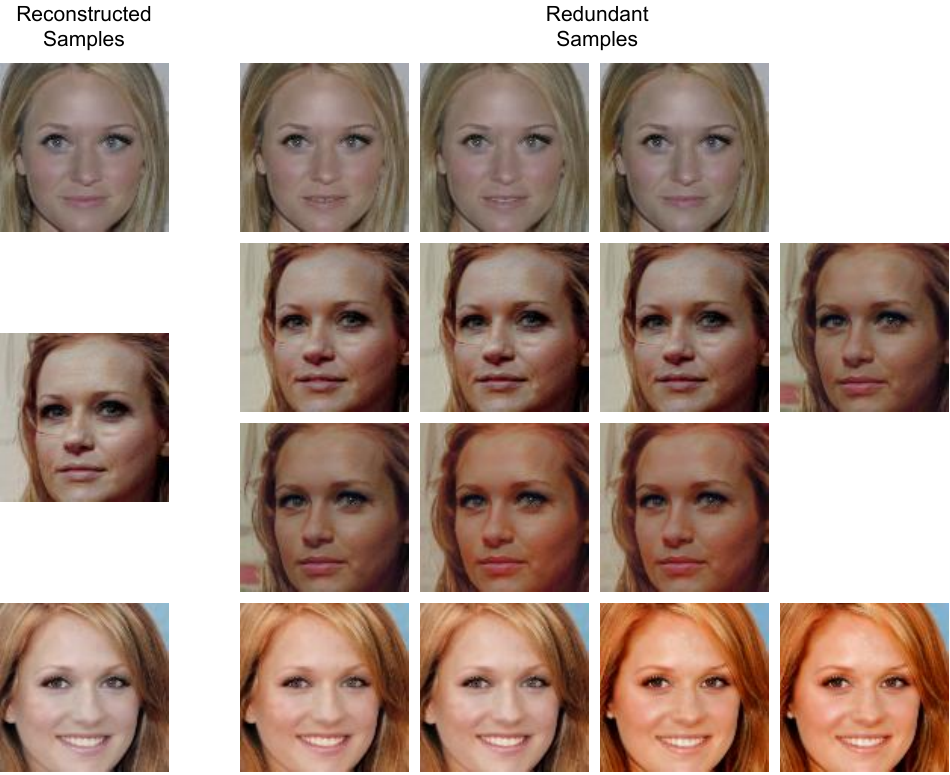}
	\caption{Several redundant samples from a target person generated with VGG16BN-UMDFaces setting.}
	%\vfill
	\label{fig:redundant_samples}
\end{figure}
Existing MI attacks optimize each reconstructed sample independently, so these samples can easily fall into similar local optima, resulting in the emergence of redundant samples.
As illustrated in Figure~\ref{fig:redundant_samples}, for each reconstructed sample, there will be a varying number of redundant samples.
These samples are very similar to each other, with only minor differences in details.
These repeated samples offer very limited effectiveness in further mining the privacy of $\mathcal{D}_{tar}$, but will instead change the distribution of $\mathcal{D}_{rec}$.

\end{document}